# Extended Abstract Version: CNN-based Human Detection System for UAVs in Search and Rescue

Nikite Mesvan

*Vel Tech - Technical University*
Chennai, India

*Abstract*—This paper proposes an approach for the task of searching and detecting human using a convolutional neural network and a Quadcopter hardware platform. A pre-trained CNN model is applied to a Raspberry Pi B and a single camera is equipped at the bottom of the Quadcopter. The Quadcopter uses accelerometer-gyroscope sensor and ultrasonic sensor for balancing control. However, these sensors are susceptible to noise caused by the driving forces such as the vibration of the motors, thus, noise processing is implemented. Experiments proved that the system works well on the Raspberry Pi B with a processing speed of 3 fps.

## I. INTRODUCTION

Nowadays, Unmanned Aerial Vehicles (UAVs) are increasingly used by individuals and organizations because of its wide applications in commercial, military, medical, and especially in search and rescue efforts. Moreover, the cost of using UAVs is much smaller than helicopters or airplanes. Currently, drones are often equipped with cameras to view the scene in real time. Flights near the disaster areas and the images obtained could help researchers assess damage and find survivors. Furthermore, UAVs can provide medical supplies to humans in quarantine or inaccessible areas. These are tasks that experts say will take much time and manpower to complete with human. Quadcopter, or drone, is a type of UAVs, which is composed of four motors, and also is the most common type of UAVs used in industry and research due to its flexibility in movement, compact size, and ease of production [1].

Currently, the search and rescue mission is studied by experts, especially in supporting victims of a disaster. A potential approach proposed in this paper is to use drone as a supporting robot because its traveling path is less blocked from the air than that of the unmanned ground vehicles (UGVs). On the other side, one of the most important parts of the mission is human detection. Researches now on object detection are still developed day by day [5,6,7,8]. YOLO and SSD are the state-of-the-art deep learning object detection methods that are demonstrating promising results and yielding faster detectors with impressive accuracy. This paper presents a solution for human detection using a deep learning neural network, Single Shot Detector (SSD), with a Quadcopter platform in the application of search and rescue.

## II. PROPOSED SYSTEM

### A. Experimental Apparatus

Figure 1 depicts the block diagram of the system. The system consists of a microcontroller Arduino Uno, 4 Electric Speed Controllers (ESCs), 4 Brushless DC motors, an Ultrasonic Sensor HC-HR04, a MPU 6050 sensor equipped with 6-degree-of-freedom inertial measurement unit: 3-axis gyroscope and 3-axis accelerometer, a Raspberry pi 3 model B, a 5.0 MP camera for the maximum resolution is 720p, and the resolution of 640x480 with 30fps at the bottom to look down vertically, and a 3800mAh battery for continuous flights from 10 to 15 minutes. The quadcopter can achieve the speed of about 3 m/s.

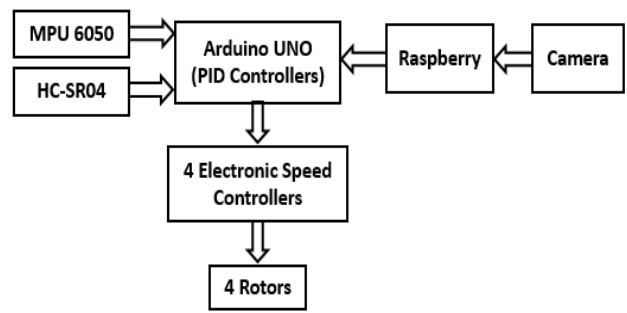

*Fig. 1. The block diagram of the system.*

### B. PID Controller

PID (Proportional-Integral-Derivative) controller is the commonest control algorithm used in many applications to optimize the system automatically. The yaw, pitch and roll angles of the quadcopter are initialized as figure 2.

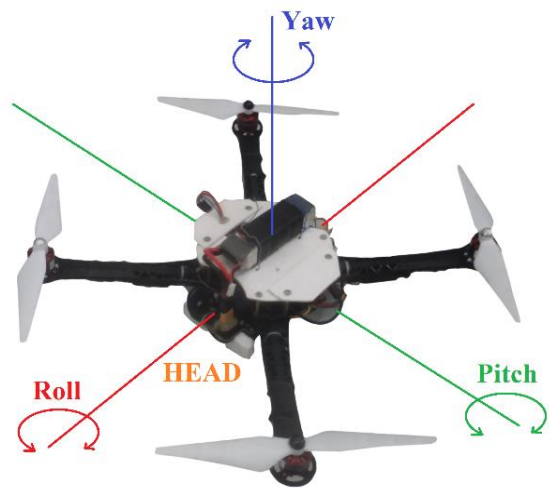

*Fig. 2. The yaw, pitch and roll angles.*

In addition, in this platform, an ultrasonic sensor is used to maintain the altitude of the quadcopter. In this case, the echo time signal from the ultrasonic sensor is fed to a PID controller [1].

Ultimately, apply the yaw, pitch, roll angles and the echo time signal to the PID controllers. The system finally is structured as in figure 3.

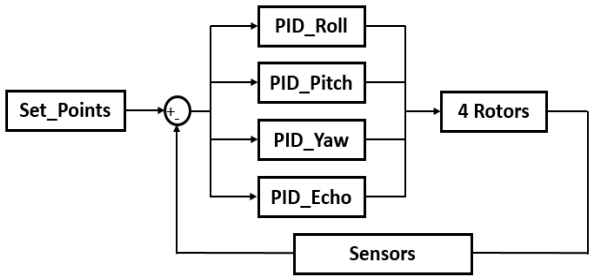

*Fig. 3. The control of the quadcopter.*

*C. Noise Processing*

An accelerometer itself is a sensor susceptible to interferences. In quadcopters, the driving forces of the system, such as the vibration of the motors, will also affect the accelerometer. Therefore, using a filter is definitely necessary [1]. Besides, because of the integration over time, the measurement of the gyroscope tends to drift and do not return to zero when the system returns to its original position. The gyro data is only reliable in a short term and it starts to drift in a long term. A complementary filter indicates an efficient solution. In the short term, the data from the gyroscope is used because of its high accuracy. In the long term, the data from the accelerometer is used since it does not drift. The filter will check whenever the values measured from the accelerometer is either reasonable or not. If any value is too large or too small, it is a complete disturbance, and the complementary filter attempts to decrease the influence of this disturbance for the better computation.

*D. Single Shot Detector (SSD)*

SSD produces a set of boxes in different sizes and scores for the confidence of the object appearance in those boxes, followed by the non-max suppression method to perform the final detection [2]. SSD only runs a convolutional neural network on the input image once and computes a feature map. Because of the operation of the convolutional layers at different scales, it is able to detect objects at various scales.

## III. EXPERIMENTS

Figure 4(a-h) depict the experiments of human detection with the pre-trained model, the results proved that the model is able to detect people at different postures and perspectives. Table 1 shows speed comparison between SSD and previous researches using the Histogram of Oriented Gradients (HOG) Detector [3], the Haar Cascade method [4], which were applied to human-searching UAVs, the experimental results demonstrated that the SSD model is better in human detection application. For further detail, SSD indicated a better performance than the HOG method and the Haar Cascade method with an average processing-speed of 3 fps (in the previous researches, the HOG Detector indicated an average computation time of 18.879 sec per image [3], and the Haar Cascade method produced a processing-speed of 1 fps [4]). Visual human detection results are showed in figure 4 and figure 5.

In difficult contexts, described in figure 5(a-d), when images may not catch fully a person into the frame of the camera, the model still produces encouraging results, which previous methods such as Haar Cascade and HOG have not worked well [3,4].

Table 1. SSD: Speed comparison in human detection for drone with the previous methods (Haar Cascade and HOG).

| Method | FPS | Sec per image |
|---|---|---|
| HOG | 0.053 | 18.879 |
| Haar Cascade | 1 | 1 |
| **SSD** | **3** | **0.333** |

## IV. CONCLUSIONS

This paper has proposed an approach for human search mission using a convolutional neural network and quadcopter. The model used in the research is a pre-trained model and also is suitable for running on the Raspberry Pi hardware platform. Experiments proved that the Quadcopter is able to flight stably and the CNN model works well on the Raspberry Pi with the processing speed of average 3 fps. In more difficult tasks, such as the images which do not display completely and fully a person into the frame of the camera, the model also demonstrates encouraging results.

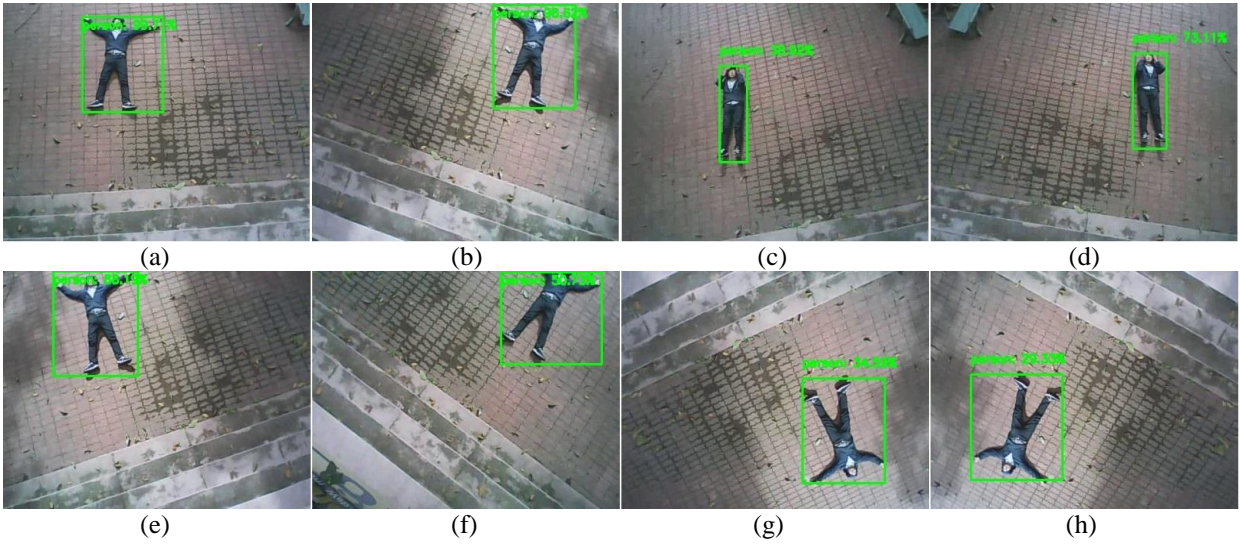

Fig. 4. Human detections at different postures and perspectives.

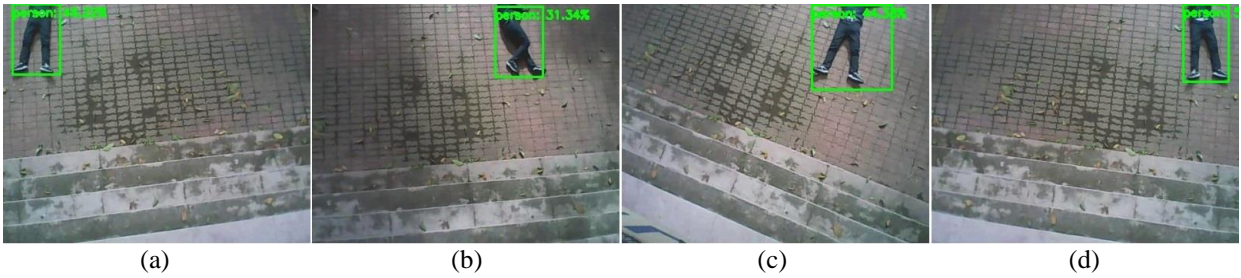

Fig. 5. Detection results in images which do not catch fully a person into the frame of the camera.